\documentclass[letterpaper,  10pt, conference]{ieeeconf}  

\IEEEoverridecommandlockouts                              

\usepackage{graphics} 
\usepackage{epsfig} 
\usepackage{mathptmx} 
\usepackage{times} 
\usepackage{amsmath} 
\usepackage{amssymb}  
\usepackage[hidelinks]{hyperref}
\usepackage[style=numeric, backend=biber]{biblatex}
\usepackage{float}
\usepackage{caption}
\usepackage{multicol}
\usepackage{cuted}
\usepackage{stfloats}
\usepackage{color}
\usepackage[nonumberlist,acronym]{glossaries}
\usepackage{textcomp}

\newcommand{\todo}[2][]{\textcolor{red}{TODO\ifx&#1&\else(#1)\fi: #2}}

\graphicspath{{../../../img/}}

\usepackage{fancyhdr}

\newcommand{\changefont}{%
    \fontsize{7}{11}\selectfont
}
\fancyhfoffset[LE]{6mm}%
\fancyhfoffset[RO]{6mm}%

\title{\LARGE \bf
Imagination-enabled Robot Perception
}

\author{Patrick Mania$^{*}$, Franklin Kenghagho Kenfack$^{*}$, Michael Neumann$^{*}$ and Michael Beetz$^{*}$%
\thanks{$^{*}$Researchers at the Institute for Artificial Intelligence (IAI) at the University of Bremen}%
}

\begin{document}
~\\
2021 \textcopyright IEEE. Personal use of this material is permitted.
Permission from IEEE must be obtained for all other uses, in
any current or future media, including reprinting/republishing
this material for advertising or promotional purposes, creating
new collective works, for resale or redistribution to servers or
lists, or reuse of any copyrighted component of this work in
other works.
\clearpage

\maketitle
\begin{strip}
	\vspace{-1.5cm}
  \centering
 \includegraphics[width=\textwidth]{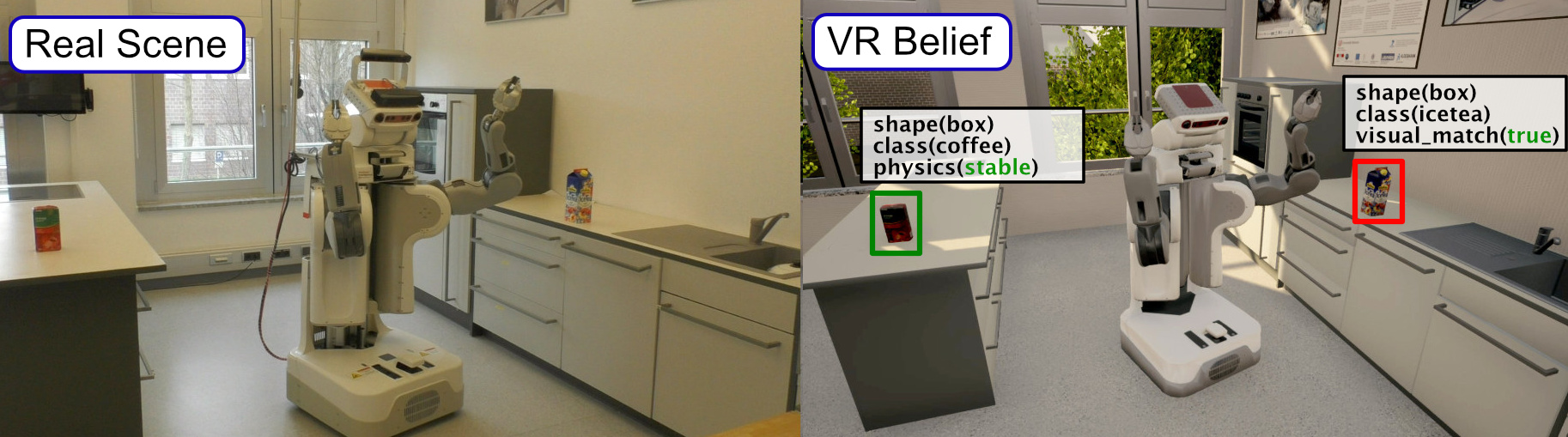}
  \captionof{figure}{Our system allows robots to ground their state, perception results and actions into a detailed, photorealistic simulation environment where every entity is linked to a knowledge base. 
		This environment acts as an expressive belief state that can be used to reason about visual expectations or refine perception predictions if the sensor data does not match the expectation.}
  \label{fig:summaryattop}
\end{strip}

\begin{abstract}
Many of today's robot perception systems aim at accomplishing perception 
tasks that are too simplistic and too hard. They are too simplistic because 
they do not require the perception systems to provide all the information 
needed to accomplish manipulation tasks. Typically the perception results 
do not include information about the part structure of objects, 
articulation mechanisms and other attributes needed for 
adapting manipulation behavior. On the other hand, the perception problems 
stated are also too hard because ---unlike humans--- the perception systems 
cannot leverage the expectations about what they will see to their full potential. 
Therefore, we investigate a variation of robot perception tasks suitable 
for robots accomplishing everyday manipulation tasks, such as household 
robots or a robot in a retail store. In such settings 
it is reasonable to assume that robots know most objects and have detailed models of them.
We propose a perception system that maintains its beliefs about 
its environment as a scene graph with physics simulation 
and visual rendering. When detecting objects, the perception system 
retrieves the model of the object and places it at the corresponding place 
in a VR-based environment model. The physics simulation ensures that object 
detections that are physically not possible are rejected and scenes can be 
rendered to generate expectations at the image level. 
The result is a perception system that can provide useful 
information for manipulation tasks.
\end{abstract}

\newacronym{systemname}{ICBS}{Imagination-capable Belief States}
\newacronym{pp}{PP}{Percept Processor}
\newacronym{aw}{AW}{Artificial World}


\section{\textbf{INTRODUCTION}}

Robots that are to accomplish manipulation tasks in human environments
need perception systems for acquiring the information needed to adapt
their behavior to the objects they are to manipulate, the tools they
can use, and the context of the scene.  Consider a household robot
that is to fetch milk and add some to the coffee. To perform this task
competently, the robot has to go to the counter, pick up the milk box,
then hold the milk box to open it by
unscrewing the lid and tilting the milk box such that the milk is
poured into the mug without spilling.

Such a robot has to be equipped with a vision system that can
``perceive'' semantic knowledge, the component and part structure of
objects, as well as their physical and geometric details as necessary 
for the execution of the task. Semantic
knowledge includes that the milk box is a container that is not
rigid. Knowledge about object-part relations includes knowledge that
the milk box has a lid that can be opened by unscrewing.
Additionally, the robot needs a tool to verify that the inferred belief 
about the scene and the relevant objects holds in the visual domain. 
For example to verify that the lid is on the detected milk by 
visualizing it internally and comparing it with the real world percepts.
This is important because robot manipulation actions highly depend 
on the quality of the belief state, 
especially in terms of the detected objects and 
their poses. 

In this case, robot perception should be viewed as a form of open
question answering based on visual information. So far, research on
robot vision systems that can do open question answering has received
surprisingly little attention. In this paper we investigate perception
systems that have these question answering capabilities for task and
environment settings where robots can make a \emph{weak closed-world}
assumption; that is that the robots know almost all 
entities in the environment. The knowledge base of the robot is
populated with object models that consist of CAD models, the
part structure, articulation models, textures, as
well as encyclopedic, common-sense and intuitive physics knowledge
about the object. We call the closed-world assumption \emph{weak}
because the robot is still required to detect novel
objects\footnote{Building detailed and realistic models for novel
  objects is outside the scope of this paper}.

The key assumption underlying our approach is that it is feasible for
a robot perception system to implement and maintain a belief state
about the environment as a scene graph in a virtual environment. Using
the rendering mechanism of the virtual environment and the knowledge
about the pose of its camera the robot can compute the image that it
expects to capture through its camera. By comparing the expected and
the captured image the robot can efficiently detect and interpret
discrepancies.

In this paper, we propose a perception system that reconstructs the
current state of the environment and the robot as an \acrfull{aw} that is
implemented using Virtual Reality(VR) technology as depicted in
Figure~\ref{fig:summaryattop}. Because this technology offers 
advanced photorealistic rendering 
and physics simulation, it allows robots to have an 
expressive method for belief state
representation and reasoning, 
that goes beyond purely symbolic approaches.

In the following, the major components of the system are discussed as well as the methods 
necessary to continously synchronize a robot perception system 
with the VR-based belief state.
Possible approaches for discrepancy detection between real world 
and expected sensor data are outlined as well as the benefits 
of our system in an exemplary object recognition and pose estimation scenario 
evaluated on a real robotic system.

\begin{figure}[t]
  \centering
  \includegraphics[width=0.93\columnwidth]{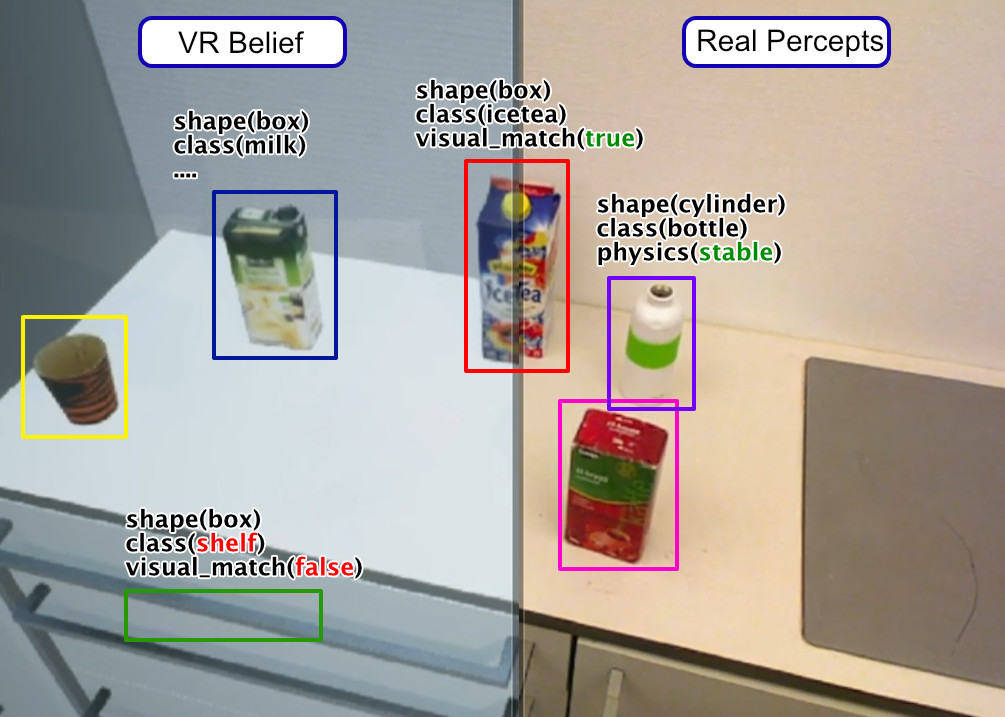}
\caption{Real world sensor data (right) next to an envisioned version of the same scene(left) from a VR engine, based on the current belief. 
	By representing the belief about objects in an VR-based artificial world, 
we can reason about visual similarity or if the prediction is physically plausable.}
  \label{fig:realvssim}
	\vspace{-\baselineskip}
\end{figure}

The main contributions of this paper are as follows:
\begin{itemize}
	\item We provide an overview of our system called \textit{\acrfull{systemname}}
		that enables robot systems to envision their belief explicitly in an \acrfull{aw} in such a way, 
		that it can be directly compared against sensor data from the real world (see Figure~\ref{fig:realvssim}). 
	\item A mechanism to initialize the \acrshort{systemname} in the \acrshort{aw} from a perception system and a real robotic system
	\item The ICBS update process used to re-recognize objects and compare the envisioned sensor data from the AW against the real 
		world sensor data to detect inplausable predictions from perception methods. 
\end{itemize}

\section{\textbf{RELATED WORK}}
In this section we want to present works that are combining VR technology with robot perception systems.

The usage of VR in the field of robotics goes back to the late 1980s 
where immerse tele-operation methods have been developed \cite{1087127}.
Even though this approach helps reducing the tele-operation costs and improving the interactivity 
between the controller and the robots, it does not lead to significantly more autonomy. 
VR has also been intensively used to train robots. 
It has been a promising environment for machine learning methods such as 
reinforcement learning(\cite{sallab2017deep},\cite{chen2017end}), 
which is not always feasible to conduct in the real world. 
The focus is often laid on specific perception tasks 
such as navigation in rather abstract geometric worlds.

A major contribution of VR to robot perception is the synthesis of big datasets(\cite{qiu2016unrealcv},\cite{muller2018sim4cv}). 
However, this data is often subject to disembodiment and unsituatedness: 
The training data does not reflect the robot's real target environment and architecture (e.g. camera properties, model of the environment, instrinsic features).
The work in \cite{mania2019framework} provides partial solutions to tackle the 
problem of the disembodiment and unsituatedness
by interfacing the control program of the robot and modeling the environment.
However, being an offline training data generation approach, 
it does not provide an online mechanism to validate perception results 
from the current robot belief against the current real world data.

Providing capabilities of mental simulation,has been examined in a few works.
A robotics-related study on mental simulation has been conducted by Cassimatis et al. in \cite{cassimatis2004integrating}. 
By developing a multi-specialist architecture based on Polyscheme they showed 
that a restricted object tracker could be improved by using symbolic high-level 
knowledge about perceived objects and their possible actions.
Most mental simulation works focused on specific tasks like navigation (\cite{rocha2019mental},\cite{yao2020path}) or localization, 
not fully exploiting all the potential simulator capabilities (e.g. realtime photorealism or physics) 
that VR potentially offers to visually verify a manipulation-focused object belief state 
by envisioning scenes. 

General mental simulation in robot perception, as indicated by the literature, is a new topic. 
An interesting approach to envision scenes are Generative Adversarial Networks(GAN), 
which can be used to generate realistic images based on semantic input 
or input images that are to be translated, like in \cite{wang2018high}.
Using GANs to realistically envision potential outcomes of robot manipulation and perception tasks 
imposes many challenges, as larger GANs tend to suffer from instabilities \cite{brock2018large},
but also need a suitable input representation for the full robot description, the task context including objects
and physical effects. By using VR we model these task-relevant semantics and phenomena effectively and can envision future states 
by executing the simulation.

Though robots have been mentally simulating their actions in artificial worlds before engaging in the real world, 
connecting mental simulation and perception systems for envisionment in manipulation tasks has only received little attention.  
Balint-Benczedi \cite{balint18iros} proposed a perception system which can instantiate 
experienced scenes from episodic memories in a VR environment to train a YOLOv3 model \cite{yolov3} on the simulated images offline.
In the past, several works have used isolated offscreen rendering techniques, 
for example as a metric for particle-based 3D-Tracking \cite{murphy2008particle}, 
pose estimation \cite{pauwels_simtrack_2015} or pointcloud-based object recognition frameworks \cite{aldoma2011cad}. 
These specialized approaches showed that information gained by reconstructing, rendering and comparing of hypotheses of the given problem is useful. 
Our approach includes these possibilities as well but also adds the potential to simulate physics phenomena 
in realtime while providing a comprehensive, photorealistic environment with annotated semantics in a broad range of domains.
To the best of our knowledge, no other works provide a similar VR-Engine-based belief state system that can operate online on a real robotic system that is targeted for manipulation actions while also providing tools to realistically envision scenes.
\begin{figure*}
  \centering
  \includegraphics[width=0.9\textwidth]{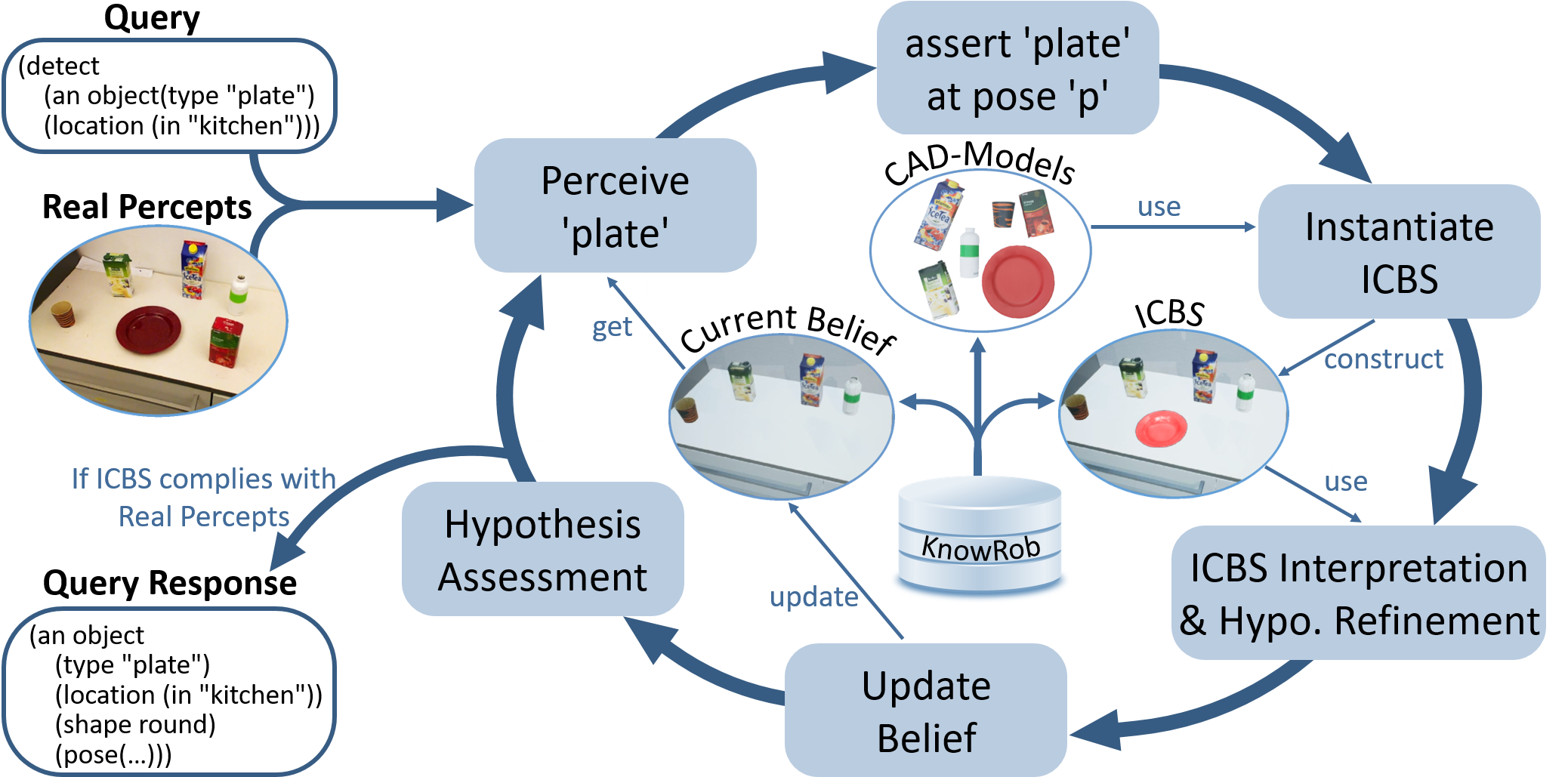}
  \caption{Our proposed architecture provides a task-based robot perception loop 
		that incorporates multiple sources of knowledge to refine predictions iteratively. 
		Passed perception tasks (queries) trigger a perception action that will instantiate 
		an artificial world with the predicted belief state (ICBS) of the robot and its environment. 
		By reasoning about the plausability of that predicted belief, 
		the robot updates its belief and either a) returns the query result to the robot control system or b) passes the current belief back to the perception action module to generate new hypotheses and analyze them in the next iteration.}
  \label{fig:architecture}
	\vspace{-\baselineskip}
\end{figure*}

\section{\textbf{ARCHITECTURE}}

In this section, we will present our developed system top-down. 
In Figure~\ref{fig:architecture}, the high-level-concept of the data flow in 
our system is visualized. Imagine that a robot is supposed to prepare 
a meal and needs to detect a plate on the table in the current task context. 
The resulting perception task is formulated as a query and passed to a perception system 
that is capable of detecting the required information in the incoming sensor data.
Based on these results, a potential belief state is constructed in the \acrshort{systemname} system by 
1) replicating the current state of the robot and 
2) asserting corresponding CAD models for detected object hypotheses into the \acrfull{aw}.
Symbolic knowledge about objects, environments and the robot has been modelled in KnowRob \cite{tenorth2013knowrob}, 
which is a knowledge representation and processing framework for robots.
The combination of this knowledge and the perception hypothesis results in the construction 
of a new, predicted belief state in the \acrshort{aw}.

In this simulated environment, a virtual camera is used to gather RGBD sensor data $S_{AW}$ from the virtual scene.
This data holds the expected visual input, given the current belief and can be compared against the incoming sensor data from the real world cameras $S_{RW}$. 
If a discrepancy between both data sources is detected, it is assumed that the initial object hypothesis was not precise enough.
The reason for this could be that a pose estimation was inaccurate or a wrong object class has been predicted.
Potential solutions to refine the belief state or matching information are stored into the belief state to keep 
the information about the plausability of the predicted belief state.
In the final step, the system decides if the envisioned belief state explains  
the real world sensor data well enough to assume that the perception hypothesis was correct. 
If this is true, the system will return the result as a response to the initial query. 
Otherwise, the process starts the next iteration of the same loop, exploiting the knowledge gained from the belief state analysis in the last iteration. 

\begin{figure}[t]
  \includegraphics[width=0.95\columnwidth]{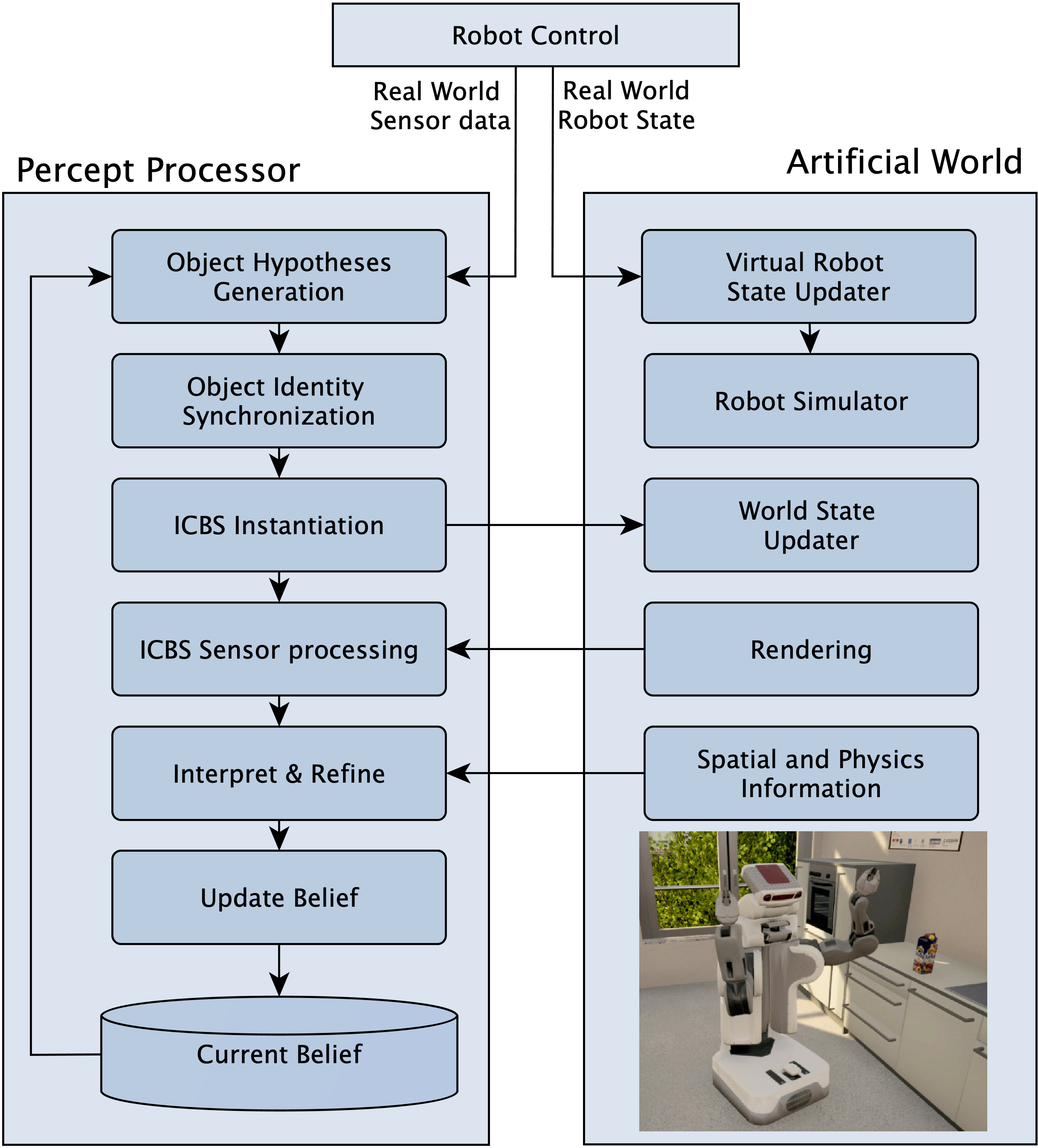}
  \caption{Interaction between the main perception functionality and the belief state in the VR-based artificial world.}
  \label{fig:architecture_small}
	\vspace{-\baselineskip}
\end{figure}

We now introduce the main components for perception and the simulation of the \acrshort{aw} in the \acrshort{systemname} system (see Figure \ref{fig:architecture_small}).

\subsection{Percept Processor} 
\label{sub:pp}
The \acrfull{pp} includes all components that interpret the incoming sensor data from the robots, executing task-related perception methods and the estimation of the current belief state. 
In our framework, we are using a perception variant of the Unstructured Information Principle \cite{ref86} which treats incoming 
sensor data $S_{RW}$   
as unstructured information that will get analyzed and annotated by multiple \textit{experts}. 
Experts can be specific methods for perception tasks 
such as shape estimation, object recognition or segmentation 
which are ultimately 
used to form object hypotheses with detected semantic properties. 
Object hypotheses are also tracked over multiple percepts by an 
Object Identity Synchronization (see Section \ref{subs:init}) 
to generate a mapping $f : o_{RW} \rightarrow o_{AW}$  between the perceived object hypotheses $o_{RW}$ and the associated virtual objects $o_{AW}$ in the \acrshort{aw}.

After the object hypothesis generation, the estimated belief of the scene is converted into commands 
to synchronize the world state in the VR-based \acrshort{aw} accordingly which includes the removal, addition or update 
of objects. 
At the same time, the real world state of the robot is replicated into the \acrshort{aw}.
The system then receives sensor data from the \acrshort{aw} denoted by $S_{AW}$ which represents 
the estimated sensor data given the estimated belief. $S_{AW}$ is a triple containing a RGB, Depth and Object Mask image.
This allows to generate 2D/3D percept data from the \acrshort{aw} that can be easily segmented. 

The current symbolic belief, the real world sensor data $S_{RW}$ and the \acrshort{aw} sensor data $S_{AW}$ 
will then be interpreted to infer if the estimated sensor data $S_{RW}$ matches the current symbolic belief. 
If the similarity criterion is not met, refinement methods such as pose refinement or additional classification experts
can be executed. Otherwise, the current belief of the perception system is updated and grounded into the knowledge 
base of the robot.

\subsection{Artificial World} 
\label{sub:aw}
The simulation technology for our \acrshort{systemname} system is Unreal Engine 4(UE4). 
It is a VR engine capable of simulating physics and  
rendering photorealistic scenery\footnote{https://www.youtube.com/watch?v=9fC20NWhx4s} 
while being able to simulate even very complex scenes in realtime. 
Being a VR engine, UE4 also offers techniques for object or environment interaction 
that are targeted for manipulation tasks. 
This allows the simulation of handling objects like unscrewing a lid or handling levers.
As it is a widely adopted game engine 
the provided features progress 
quickly. Simulating robotics-related tasks was previously mostly done 
with dedicated simulation environments like Gazebo, 
having limited support for large scale scenes and photorealism.

To use UE4 as a robot belief state, multiple components had to be developed. 
Robots are complex systems which can be described in the URDF format. 
Our system processes these descriptions to instantiate the employed robotic system in the \acrshort{aw}, 
including physics constraints to model the joints in the VR engine.
The robot state (joint states + localization) is replicated online into the \acrshort{aw}.\footnote{Sensor data is also interpreted online in the \acrshort{pp}.}
This allows the generation of embodied sensor data from the estimated belief. 

The environment can be manipulated with a ROS-compatible RPC interface that allows to add, modify or delete 
objects in the \acrshort{aw}. This is used by the \acrshort{pp}, to set up the estimated belief in the \acrshort{aw}.
All objects are handled by a physics simulation, which allows us 
to reason about object collisions in estimated object poses or if objects moved significantly after an event.
Spawn collisions are detected by the automatic construction of overlap collision boxes for every $o_{AW}$.
Physics events are recorded and related to each of the objects in the belief. 
We aggregate object-related information in a module for spatial and physics information and feed it back to the \acrshort{pp} 
for the interpretation and refinement step. Since UE4 does not support ROS directly, the communication with the other components in our system is done through rosbridge \cite{crick2017rosbridge}.

\section{\textbf{SYNCHRONIZATION OF ICBS}}
We will now outline the creation process of \acrshort{systemname}. 
The description will be split in two parts: 
1) The initialization procedure of the whole system to explain the data flow from a perception task to a 
populated \acrshort{aw}, representing the current belief and 
2) the update process that compares the sensor data $S_{AW}$ and $S_{RW}$ 
to detect mismatching expectations and leads to the correction of the current belief about the scene.

\subsection{Initialization}
\label{subs:init}
The initialization begins with a semantic map of the environment. 
This data structure describes 
the geometry and spatial relation of entities in the target space to act in. 
It is a fundamental prior for the perception system, but also for the \acrshort{aw} that is constructed in the VR engine. 
We assume that 
localization information is available. 
The robot pose will be constantly transformed into the coordinate system of the AW to update 
the virtual robot model in the AW, ultimately resulting in similar viewpoints of the current scene (see Figure~\ref{fig:realvssim}).

\begin{figure}[t]
  \centering
  \includegraphics[scale=0.5]{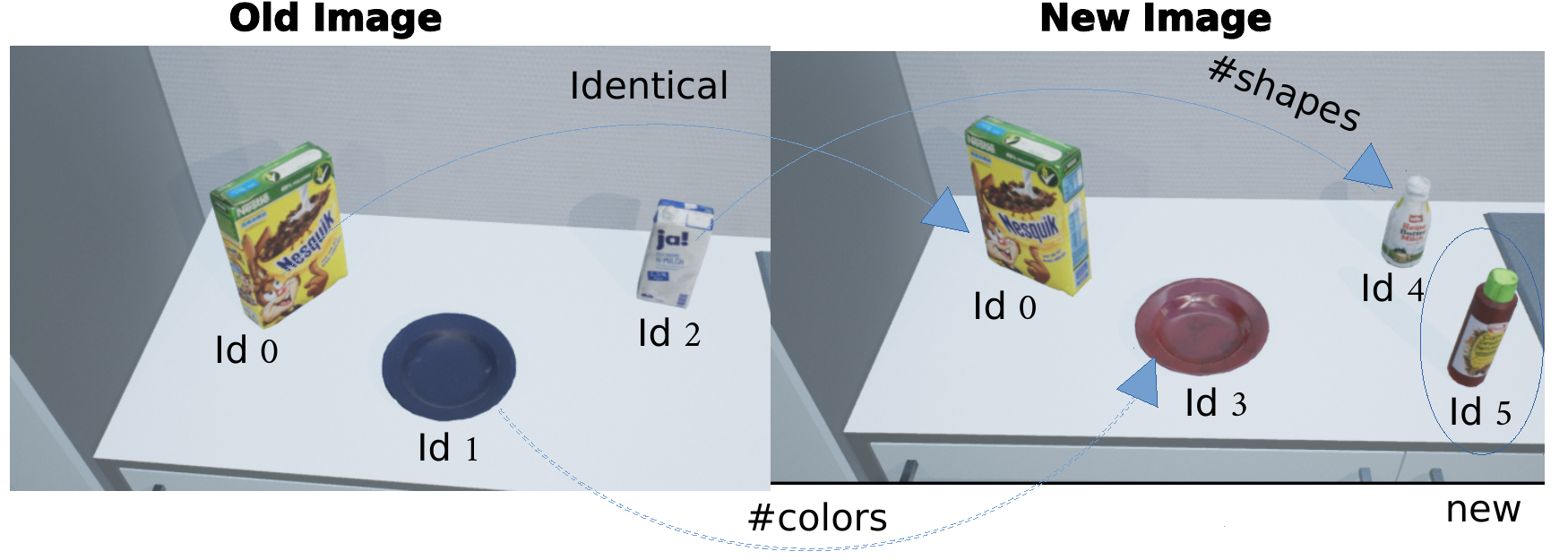}
  \caption{The Object Identity Synchronization mechanism uniquely identifies objects over time to link them in the \acrshort{systemname}. Object IDs only change if objects have been exchanged and/or are added to the scene.}
  \label{fig:ois}
	\vspace{-0.3cm}
\end{figure}

To create an initial belief, the \acrshort{pp} will segment the incoming sensor data $S_{RW}$ based on spatial information 
for relevant surfaces in the semantic map. After the segmentation, object candidates $o_{RW}$ are detected and asserted in a scene graph.
Every $o_{RW}$ is annotated with the results of the perception experts, 
like shape, color, 6D pose or predicted class labels.
An important task of the perception system is to create a consistent 
graph of the scene over time 
despite changes that might happen on the objects, for example removal, moving or adding. 
Every object in the scene graph should be re-recognized over multiple percepts 
to be able to 
reliably link the corresponding object in the \acrshort{systemname}. 
Otherwise, objects might be re-added to the belief because the perception system could not 
infer that it just saw the same object again.
To solve this problem, we implemented an Object Identity Synchronization(OIS) that uniquely identifies 
detected objects (see Figure \ref{fig:ois}) and connects them with \acrshort{systemname} entities. 
In the initialization phase, 
every $o_{RW}$ gets an unique identifier with a timestamp. 
In every following perception step, the current object hypotheses are matched against 
the known objects in the scene graph. The object identity is determined by 
a weighted policy that computes similarity measures on the  
properties 
that have been annotated by the perception experts. 
Each object in the scene graph is then linked to an entity in the \acrshort{systemname}
by predicting the object class and instantiating the particular CAD object model 
in the AW with the estimated pose, effectively recreating the 
hypothesis about the observed scene as a virtual copy. 
The corresponding virtual object for an object hypothesis $o_{RW}$ is denoted as $o_{AW}$.
Because \acrshort{systemname} are a three-dimensional 
AW, the perception system must provide 6D poses for objects.
In our scenario, we use RGBD sensors on our robot to segment and estimate the objects' pose 
in pointcloud data. 
The pose will be transformed from the frame of $S_{RW}$ into the coordinate system of 
the AW before the creation of the object $o_{AW}$ is triggered on the corresponding 
interface of the VR engine.

\subsection{Update}
After the initial state of the \acrshort{systemname} has been set up, the system must ensure 
that the belief state is always updated with the information from the incoming sensor data $S_{RW}$. 
Predictions on detected objects should be able to be related 
to their virtual counterparts $o_{AW}$ to be able to verify their plausability or trigger 
a reparametrization of the predictive models to generate new predictions.

\begin{figure}[t]
  \centering
  \includegraphics[width=\columnwidth]{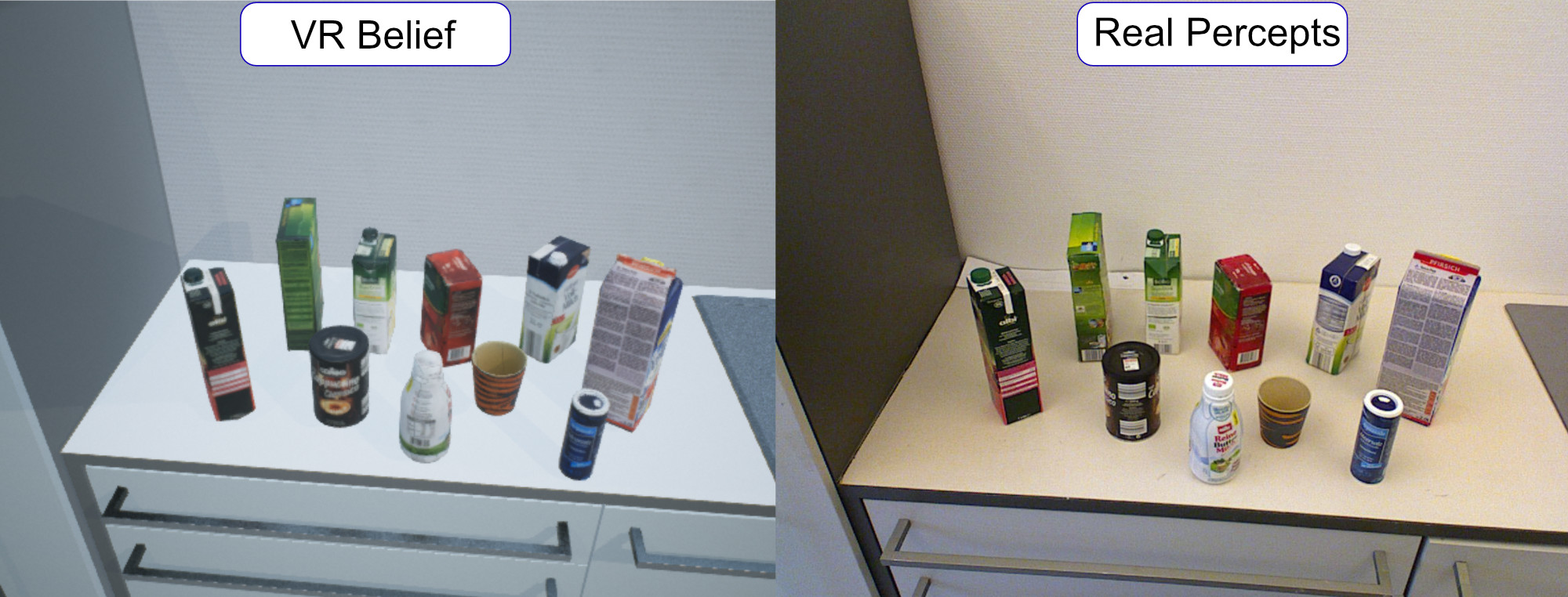}
  \caption{Resultant AW(left) after ICBS initialization and update with real world percepts(right) from a robot.}
  \label{fig:setup}
	\vspace{-0.5cm}
\end{figure}

The update process begins with the execution of the perception pipeline in the \acrshort{pp} where 
the current scene will be analyzed by the perception experts. 
In contrast to the initialization 
step, the system can now use the information in the scene graph to identify and annotate new information to 
objects that have already been discovered by the OIS in $S_{RW}$. 
Because the OIS monitors the appearance, movement or disappearance of objects, 
the update of the AW can be triggered whenever such an event is discovered. 
After the AW contains the current expectation of the appearance of the scene, 
the sensor data $S_{AW}$ and $S_{RW}$ will be compared in every region that contains 
object hypotheses.
In this step we assume, that due to the realistic rendering possibilities of the AW, 
the sensor data (and especially the RGB image) is realistic enough that even 
non-complex similarity models can be used. 
The viewpoints of $S_{RW}$ and $S_{AW}$ are also similar 
because the robot sensor pose is constantly mapped from the real into the artificial world during execution(see Figure \ref{fig:setup}). 

The employed similarity methods depend on the given perception task, formulated as a query $q$.
This is especially important in the context of robot manipulation, 
since perception tasks in that context are diverse.
Take for example the milk pouring scenario that has been outlined in the introduction. 
The robot should localize a milk box as well as validate his object recognition 
results to reduce the chance of a misdetection. 
In this scenario, we propose two comparison methods that are suitable to 
confirm the results: 1) A image-based method that compares the ROI in $S_{AW}$ and $S_{RW}$ 
depicting the detected object and 2) a physics-based validation scheme.
The latter can be implemented by using the physics simulation of the VR engine. 
Imagine for example that a picture of the wanted milk box has been detected on an advertisement 
hanging on the wall. If this object hypothesis is spawned in the \acrshort{aw}, 
the milk box will either collide with the wall or fall down on the next supporting surface.
Both cases are detected by our Spatial and Physics Information module (see Figure \ref{fig:architecture_small} and Chapter~\ref{sub:aw}) and indicate that the object recognition produced a false result.

Method 1 builds upon the idea that if one presents an image of the detected object 
in the scene and an image of the same object 
as a textured CAD model from the same viewpoint in the \acrshort{aw}, 
it should be possible to decide if both images are showing the same object. 
Our system includes several comparison methods to check for the similarity of 
feature annotations made by the experts in the \acrshort{pp} on the sensor data 
$S_{RW}$ and $S_{AW}$.
One image-based comparison method we propose are color histogram annotations 
in conjunction with the computation of their Hellinger distance as a similarity measure. 
In this step, many additional methods could be used to compare both sensor dates, 
for example edge-based contour comparisons as well as keypoint feature matching in images 
or pointcloud-based model-fitting approaches.

The capabilities of the VR engine allow to represent fine-grained object models in conjunction with 
articulation models. For example, the lid of the aforementioned milk box can also 
be modelled in VR with a constraint-based physics model \cite{emperore2015unreal} for screwing.
Robots could then simulate planned actions while having 
ground truth for all relevant entities in the AW. 
After the planned action has been simulated, the visual expectation of the scene 
can be compared to the appearance of the scene after the robot has executed the same task in the real world.
Depth-based methods can be used to compare if the manipulated objects or 
parts of the robot are now in the envisioned state (for example, 
a visible opening in the milk box where the lid was after unscrewing) 
and use this as a success criterion for manipulation tasks.
This paper focuses on providing a foundation for this functionality by proposing a method 
to set up and synchronize the typically complex robot belief states in a VR engine 
while also presenting means of verifying these belief states based on the perception results of the robot.

\section{\textbf{EXPERIMENTS \& DISCUSSION}}
\begin{figure}[t]
  \centering
  \includegraphics[width=\columnwidth]{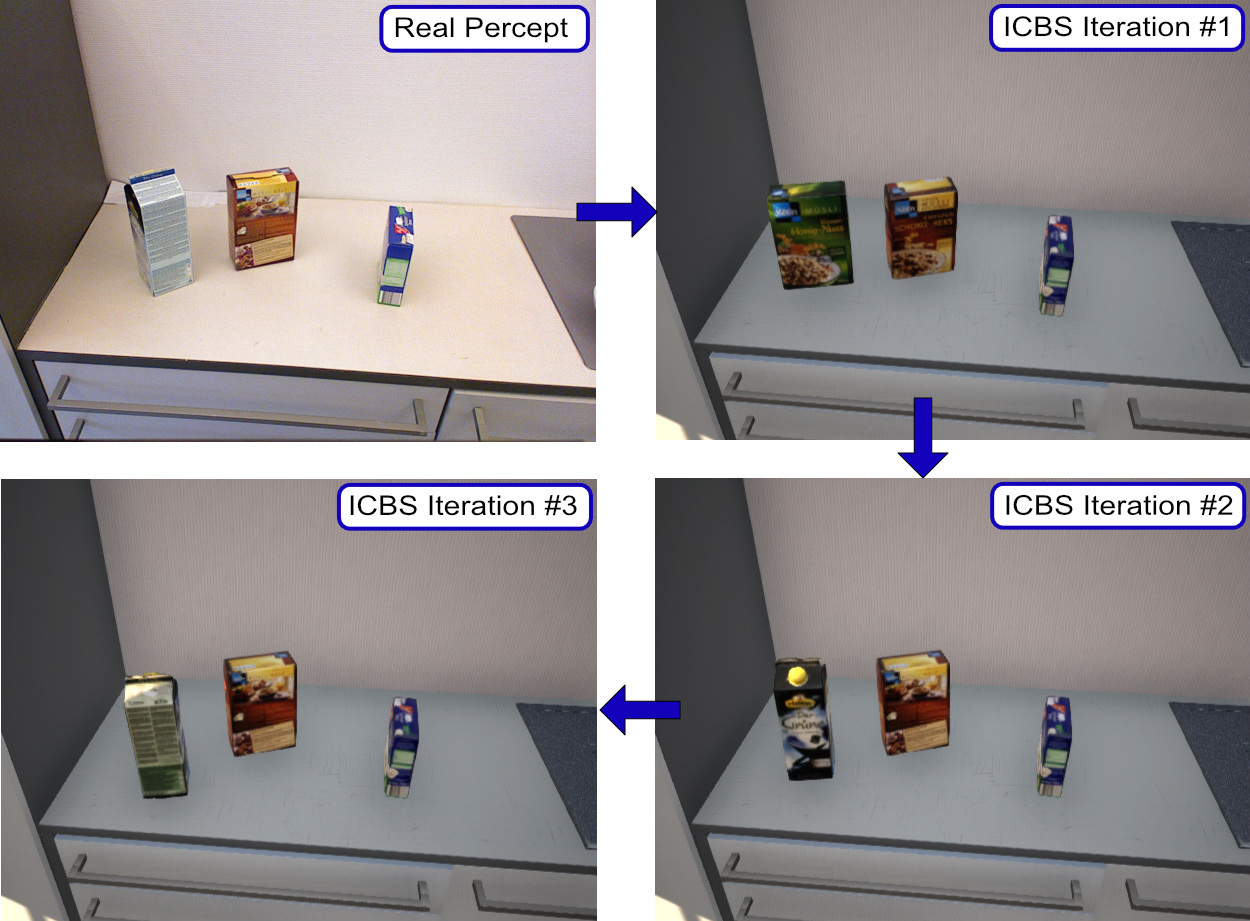}
  \caption{Example from our experiments. Based on a Real World RGBD sensor input (upper left), we reconstruct the robots' belief state in the VR engine. 
		After generating synthetic sensor data from this envisioned belief, the appearance of detected objects will be compared against their virtual counterparts to detect misclassifications or large pose deviations (see Iteration 1, upper right). The belief state can be iteratively varied (Iteration 2 and 3) and continously visually compared to improve it.}
  \label{fig:verification-evolution}
	\vspace{-0.5cm}
\end{figure}

The proposed system should support mobile robotic agents to reason 
about their envisioned belief state and planned actions in manifold ways in unstructured environments. 
Since manipulation actions and their intended outcomes are strongly dependent on valid perception results, 
we focus our evaluation in this work on an experiment for object recognition verification as well as pose estimation.
We want to emphasize that the selection of the chosen 
computer vision approaches and their parameters are not crucial to use our framework, 
as it is general enough to exchange these methods arbitrarily.

In this work, we want to investigate the potential of our system with an image-based refinement method.
The experiments have been conducted in the kitchen lab of the Insitute for Artificial Intelligence. 
A PR2 mobile robot platform \cite{bohren2011towards} is using the \acrshort{systemname} system to detect objects 
on surface regions in the kitchen where objects are located. 
The segmentation is based on the semantic knowledge about the environment and 
the surfaces of interest as well as a 3D-based clustering approach. 
Object recognition is done by using a CNN as a feature generator \cite{pmlr-v32-donahue14} 
combined with a k-NN classification model. 
The output of these methods is then used to envision the belief state in the AW which is a virtual model of our kitchen lab.
We then analyze which parts of the virtual sensor data comply with the real world sensor data.
To achieve the latter, we implemented a RGB-based appearance comparison based on 
distance measures for color histograms on segmented objects in $S_{RW}$ and $S_{AW}$.
In our experiments, we use the Hellinger distance defined as:
\begin{equation}
d(H_1,H_2) = 	\sqrt{1 - \frac{1}{\sqrt{\hat{H_1}\hat{H_2}N^2}} \sum_{I}\sqrt{H_1(I)H_2(I)} }
\end{equation}
where $\hat{H_x}$ is the average bin value and $N$ is the total number of histogram bins.
If the distance exceeds the matching threshold, 
it is assumed that the object recognition predicted a false hypothesis which will then be rejected. 

\begin{figure}[t]
\centering
\includegraphics[width=\columnwidth]{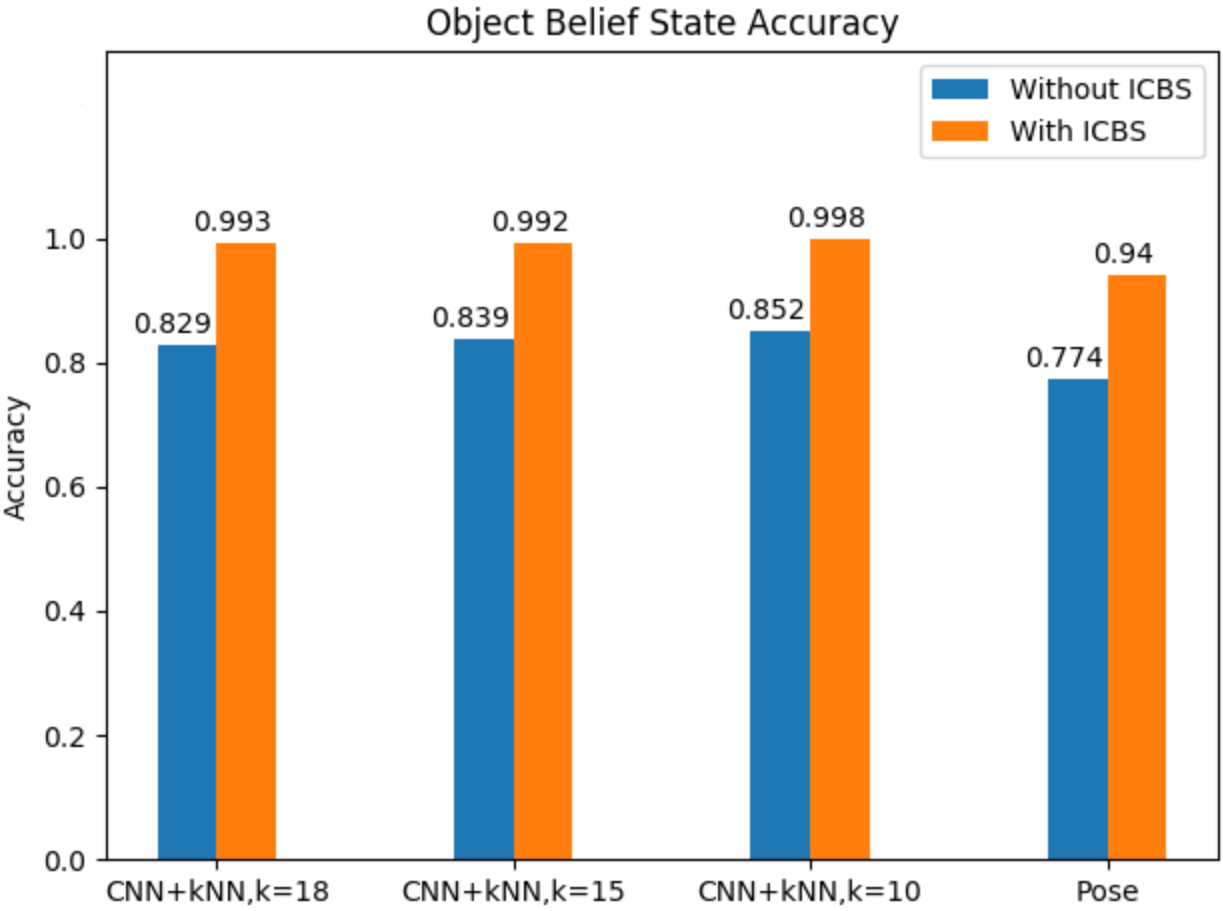}
\caption{Accuracy for experiments with and without the usage of our framework ICBS. We studied the effect of our framework on classification tasks and pose estimation.}
\label{fig:experiments}
\vspace{-0.5cm}
\end{figure}

Our dataset consists of 275 scene percepts with a total of 822 object recognition predictions.
In our experiments we employ different parametrizations of the k-NN classifier that is used to 
classify the extracted CNN features. Our goal is not to propose a certain perception approach or parametrization,
but rather show that our system provides a method for validating and improving perception results by comparing 
the envisioned scene with the real world sensor data.

With varying $k$ in the k-NN classifier, the classification performance ranges between an accuracy of 0.829 and 0.852.
By applying our ICBS system with an image-based similarity measure, we were able to envision the 
predictions visually and reject large parts of the 
false predictions, effectively increasing the accuracy (see Figure~\ref{fig:experiments}).
In some cases, objects were confused with a different variant of the same product category 
(for example, an icetea box with two different flavors from the same brand) 
which could not be detected by the employed comparison method. 
As the false predictions were made on the rather featureless backside of the presented products,
we are optimistic that this problem can be solved in future work by comparing more detailed features 
like texts.

In our second experiment, we studied a pose estimation problem. 
When analyzing the distribution of errors of computed poses, 
available methods might feature a local maxima around 180 degrees \cite{ozuysal2009pose}.
In a household or in a general pick and place scenario, 
this error results in manipulation actions where robots can not reliably handle objects where it is necessary to distinguish and plan actions based on the correct front-to-back orientation of the object.
This problem motivates our second experiment, where we visually compare, analogous to our first experiment, the appearance of $S_{RW}$ and $S_{AW}$ on varying poses of objects $o_{AW}$ to detect poses that are skewed by 180 degrees around the up-axis of the object.
The employed pose estimation method is a PCA-based geometry and pose estimator based on \cite{ref86}.
As shown in Figure~\ref{fig:experiments}, we could succesfully identify occurences of skewed poses and improve them.
Errors where usually prevalent on objects which featured a very similar front and back side which made a differentation hard.

\section{\textbf{CONCLUSION\& FUTURE WORK}}
We presented an integrated system that allows robots to form 
expectations and validate beliefs about their environment in a VR-based world model,
which is capable of simulating physics as well as rendering 
realistic imagery. By maintaining a scene graph representation 
over time, detected entities can be uniquely identified and linked to 
virtual counterparts in the VR engine.
Since everything in VR is machine-understandable,
robot control programs can make use of this feature-rich 
belief state representation to validate their perception results 
or could imagine how the environment would look like after simulating 
their planned action in the artificial world. This method of representing 
belief states allows us to expand purely symbolic knowledge representations,
effectively allowing robots to reason about subsymbolic effects in the world.
In our experiment we showcased the potential of our proposed system 
in an object recognition and pose estimation study by comparing the visual expectation of 
the predicted scene with the real world data.

Since our proposed system provides a general way of representing and 
comparing expectations about belief states, 
many potential research topics are open for future work.
In computer graphics interesting approaches for the simulation of non-rigid 
and evolving objects such as bakery products \cite{10.1145/3355089.3356537} have been developed 
that could help household robots to understand the process of making food. 
Integrating such a simulation into our system could help us to build 
better perception models that could infer in which 
state a certain type of food currently is and reason about the 
potential next steps for it (,,put in oven''? ,,serve to guest''?).
Another potential topic is the integration of robot high-level 
control into the VR engine. This would allow robots to imagine 
complex manipulation activities in an artificial world while having 
visual expectations for every action available, allowing robots 
to detect failures in the real world more easily.

\section*{\textbf{ACKNOWLEDGEMENT}}
The research reported in this paper has been (partially) supported by the BMBF project RoPHa (FKZ: 16SV7835K) and by the German Research Foundation DFG, as part of Collaborative Research Center (Sonderforschungsbereich) 1320 “EASE - Everyday Activity Science and Engineering”, University of Bremen (http://www.ease-crc.org/). The research was conducted in subproject R03 Embodied simulation-enabled reasoning.

\addtolength{\textheight}{-5cm}   

\printbibliography[title={REFERENCES}]
\end{document}